\documentclass{article}

\PassOptionsToPackage{numbers, compress}{natbib}

    \usepackage[preprint]{neurips_2025}

\usepackage[utf8]{inputenc} 
\usepackage[T1]{fontenc}    
\usepackage{hyperref}       
\usepackage{url}            
\usepackage{booktabs}       
\usepackage{amsfonts}       
\usepackage{nicefrac}       
\usepackage{microtype}      
\usepackage{xcolor}         

\usepackage[permil]{overpic}
\usepackage{epsfig}
\usepackage{graphicx}
\usepackage{amsmath}
\usepackage{amssymb}

\usepackage{colortbl}       

\usepackage{lipsum}
\usepackage{todonotes}
\usepackage{makecell}

\usepackage{blindtext}
\usepackage{ctable}

\usepackage{multirow}
\usepackage{float}
\usepackage{bbm}
\usepackage{comment}
\usepackage{mathrsfs}

\usepackage{wrapfig}

\def\eg{\emph{e.g}.} 

\def\ie{\emph{i.e}.}

\usepackage{bm}

\newcommand{\phil}[1]{#1}
\newcommand{\rsz}[1]{#1}
\newcommand{\rszfinal}[1]{#1}

\newcommand{\ed}[1]{#1}

\newcommand{\ourmethod}{COS3D}

\title{COS3D: Collaborative Open-Vocabulary\\ 3D Segmentation}

\author{
Runsong Zhu$^1$ \quad
Ka-Hei Hui$^2$ \quad
Zhengzhe Liu$^3$ \quad 
Qianyi Wu$^4$ \quad
Weiliang Tang$^1$ \quad
\\
\textbf{Shi Qiu}$^1$ \quad
\textbf{Pheng-Ann Heng}$^1$ \quad 
\textbf{Chi-Wing Fu}$^1$
\\
$^1$ The Chinese University of Hong Kong  \quad $^2$ Autodesk AI Lab \\
$^3$ Lingnan University \quad $^4$ Monash University}

\definecolor{ao}{rgb}{0.0, 0.0, 1.0}
\definecolor{airforceblue}{rgb}{0.36, 0.54, 0.66}
\definecolor{ceruleanblue}{rgb}{0.16, 0.32, 0.75}
\definecolor{cerulean}{rgb}{0.0, 0.48, 0.65}
\definecolor{celestialblue}{rgb}{0.29, 0.59, 0.82}
\definecolor{azure(colorwheel)}{rgb}{0.0, 0.5, 1.0}
\definecolor{babyblue}{rgb}{0.54, 0.81, 0.94}
\definecolor{babyblueeyes}{rgb}{0.63, 0.79, 0.95}
\definecolor{ballblue}{rgb}{0.13, 0.67, 0.8}

\definecolor{asparagus}{rgb}{0.53, 0.66, 0.42}
\definecolor{ao(english)}{rgb}{0.0, 0.5, 0.0}
\definecolor{applegreen}{rgb}{0.55, 0.71, 0.0}
\definecolor{armygreen}{rgb}{0.29, 0.33, 0.13}
\definecolor{gray-asparagus}{rgb}{0.27, 0.35, 0.27}
\definecolor{green(ryb)}{rgb}{0.4, 0.69, 0.2}

\definecolor{amethyst}{rgb}{0.6, 0.4, 0.8}
\definecolor{antiquefuchsia}{rgb}{0.57, 0.36, 0.51}
\definecolor{blue-violet}{rgb}{0.54, 0.17, 0.89}
\definecolor{brightlavender}{rgb}{0.75, 0.58, 0.89}
\definecolor{brightube}{rgb}{0.82, 0.62, 0.91}
\definecolor{brilliantlavender}{rgb}{0.96, 0.73, 1.0}

\definecolor{amber}{rgb}{1.0, 0.75, 0.0}
\definecolor{amber(sae/ece)}{rgb}{1.0, 0.49, 0.0}
\definecolor{atomictangerine}{rgb}{1.0, 0.6, 0.4}
\definecolor{burntorange}{rgb}{0.8, 0.33, 0.0}
\definecolor{burntsienna}{rgb}{0.91, 0.45, 0.32}
\definecolor{cadmiumorange}{rgb}{0.93, 0.53, 0.18}
\definecolor{carrotorange}{rgb}{0.93, 0.57, 0.13}
\definecolor{chocolate(web)}{rgb}{0.82, 0.41, 0.12}
\definecolor{chromeyellow}{rgb}{1.0, 0.65, 0.0}
\definecolor{darkorange}{rgb}{1.0, 0.55, 0.0}
\definecolor{darktangerine}{rgb}{1.0, 0.66, 0.07}
\definecolor{deepcarrotorange}{rgb}{0.91, 0.41, 0.17}
\definecolor{deepsaffron}{rgb}{1.0, 0.6, 0.2}
\definecolor{fulvous}{rgb}{0.86, 0.52, 0.0}

\begin{document}

\maketitle

\begin{abstract}
Open-vocabulary 3D segmentation is a fundamental yet challenging task, requiring a mutual understanding of both segmentation and language.
However, existing Gaussian-splatting-based methods rely either on a single 3D language field, leading to inferior segmentation, or on pre-computed class-agnostic segmentations, suffering from error accumulation.
To address these limitations, we present \ourmethod{}, a new collaborative prompt-segmentation framework that contributes to effectively integrating complementary language and segmentation cues throughout its entire pipeline.
We first introduce the new concept of collaborative field, comprising an instance field and a language field, as the cornerstone for collaboration.
During training, to effectively construct the collaborative field, our key idea is to capture the intrinsic relationship between the instance field and language field, through a novel instance-to-language feature mapping and designing an efficient two-stage training strategy.
During inference, to bridge distinct characteristics of the two fields, we further design an adaptive language-to-instance prompt refinement, promoting high-quality prompt-segmentation inference.
Extensive experiments not only demonstrate \ourmethod{}'s leading performance over existing methods on two widely-used benchmarks but also show its high potential to various applications,~\ie, novel image-based 3D segmentation, hierarchical segmentation, and robotics. 
The code is publicly available at \href{https://github.com/Runsong123/COS3D}{https://github.com/Runsong123/COS3D}.
\end{abstract} 
\section{Introduction}
\label{sec:intro}
Open-vocabulary 3D segmentation (OV3DS) aims to predict 3D segmentation of scenes according to given natural language queries.
Beyond traditional 3D segmentation, which is often restricted to fixed object categories~\cite{qi2017pointnet, qi2017pointnet++, qian2022pointnext, robert2023efficient, riegler2017octnet, siddiqui2023panoptic, bhalgat2023contrastive, zhu2024pcf}, the OV3DS task supports flexible text queries, allowing for diverse semantic categories, physical properties, affordance, and more.
This flexibility is crucial to making OV3DS a practical and valuable tool for applications in fields such as AR, VR, and robotics.

\begin{figure}[!t]
\centering
\begin{overpic}[width=0.95\textwidth]{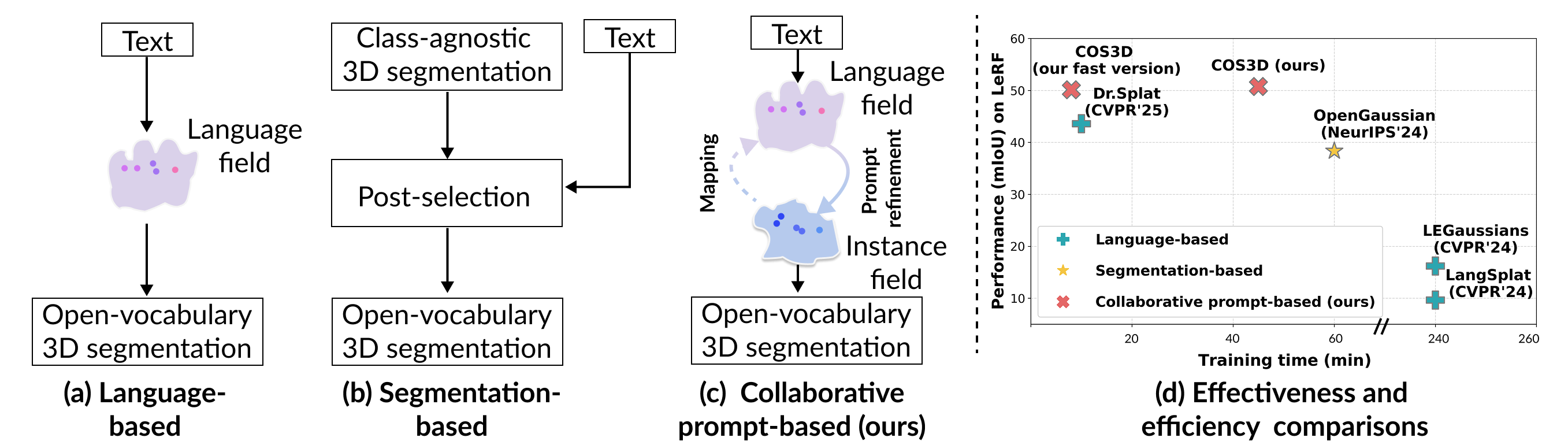}
\end{overpic}
\centering
\vspace{-2.0mm}
\caption{ 
Comparing different paradigms.
(a) \rsz{Language-based methods}~\cite{kerr2023lerf, qin2024langsplat, Shi_2024_CVPR, jun2025dr, ji2025fastlgs} directly learn a 3D language field for open-vocabulary segmentation.
(b) \rsz{Segmentation-based methods}~\cite{ye2023gaussian, wu2024opengaussian, li2024instancegaussian} perform class-agnostic segmentation then post selection.
%
(c) 
\ourmethod{} addresses existing limitations with a new collaborative prompt-segmentation framework that connects knowledge from the language and instance fields in the training and inference.
The solid line (dotted line) indicates inference (training).
(d) \ourmethod{} clearly outperforms existing methods on  both segmentation quality and training efficiency. 
Note that ``fast version'' refers to a result of our approach under a short optimization time setting (see Sec.\ref{sec:ablation}), while the time and performance of the baselines are sourced from their  publications~\cite{jun2025dr,wu2024opengaussian}.
}
\vspace{-5.5mm}
\label{fig:teaser}
\end{figure}

Recent efforts focus on transferring 2D vision-language models (VLMs) to 3D scenes represented by \phil{learned} radiance fields.
These works can be roughly divided into two classes: {\em language-based\/} and {\em segmentation-based\/}.
Specifically, language-based methods~\cite{kerr2023lerf, qin2024langsplat, Shi_2024_CVPR, jun2025dr, ji2025fastlgs} propose to distill language features (\eg, CLIP~\cite{clip_method}) from the 2D image space to a 3D language field by leveraging differentiable rendering to support OV3DS; see Fig.~\ref{fig:teaser} (a).
\rsz{However, directly learning language features through \phil{a} pixel-wise language distillation demonstrates limited distinctiveness, leading to severe artifacts and errors around boundaries in the segmentation results.
}

\rsz{On the other hand, segmentation-based methods,~\eg,~\cite{ye2023gaussian, wu2024opengaussian, li2024instancegaussian}}, directly decompose the OV3DS task into two sub-tasks: (i) a class-agnostic 3D segmentation, followed by (ii) a post-selection of the best-matched 3D segments, using a 2D vision-language model~\cite{clip_method, liu2024grounding}; see Fig.~\ref{fig:teaser} (b).
Though this approach bypasses direct language distillation, it faces two major challenges, leading to limited performance.
First, without semantic cues, accurately segmenting all objects in a 3D scene is highly challenging, so under- and over-segmentation errors often occur in the class-agnostic segmentation, which further affects the post selection.  
Second, the hand-crafted matching strategies in the post selection easily introduce additional inaccuracies that further degrade the performance.

%
Revisiting the existing methods, we attribute their limitations to the lack of integrating language and segmentation information.
In particular, these two types of information \phil{provide} complementary knowledge: segmentation information is typically \textit{discriminative} and \textit{boundary-aware}, \phil{whereas} language information facilitates high-level 
\phil{\textit{understanding of objects and scenes}}.
Fundamentally, \phil{to achieve OV3DS} requires a mutual understanding of {\em both\/} language and segmentation.

To this end, we introduce \phil{COS3D}, a new COllaborative approach for prompt-Segmentation of 3D scenes, in which we collectively incorporate segmentation and language cues in our framework; see Fig.~\ref{fig:teaser} (c).
\phil{There are three technical components in \ourmethod{}.}
First, we propose the new concept of \textit{collaborative field}, comprising an instance field and a language field, as the foundation in \ourmethod{}.
To effectively construct \phil{the two fields}, \textit{our key insight lies in their intrinsic relationship: regions within the same object segment \phil{should} share 
\phil{similar} semantics and exhibit similar language information.}
\phil{Second,} we propose modeling the intrinsic relationship through a feature mapping process from a learned, boundary-aware instance field to the text-aligned language field.
\phil{Here,} we first train the instance field to implicitly encode the segmentation information, then \phil{formulate an} instance-to-language mapping learning to facilitate the \phil{language-field} construction.
\phil{Third, } at inference, given a text query, we generate the segmentation from the text-aligned language field.
\phil{Importantly, considering the limited expressivity of the language feature}, we leverage the distinct characteristics of the instance field and introduce an adaptive language-to-instance prompt refinement to exploit the intermediate 3D relevance map from the instance field as a prompt, \phil{then design a further refinement} on the boundary-aware instance field for prompt segmentation.
With these \phil{new} designs, \ourmethod{} \phil{is able to} arrive at a surprisingly \textbf{effective} and \textbf{efficient} solution; see Fig.~\ref{fig:teaser} (d).

We evaluate \ourmethod{} on two standard benchmarks for OV3DS.
Both quantitative and qualitative results show that our method {\em significantly\/} outperforms existing approaches.  
Also, the ablation studies validate the effectiveness of our designs for both training and inference stages.
Furthermore, we present three example applications of \ourmethod{}, including novel image-based 3D segmentation, explicit hierarchical OV3DS, and robotics, demonstrating its potential and practical values.

Our major contributions are summarized as follows:

\vspace{-2mm}
\begin{itemize}
\setlength\itemsep{0.1em}
\item We present \ourmethod{}, a \phil{new} collaborative prompt-segmentation framework that integrates segmentation and language cues, 
enabling 
\phil{top}-quality open-vocabulary 3D segmentation.

\item For training, we propose a novel instance-to-language mapping with two optional implementations that effectively and efficiently leverage the instance field to enhance the construction of a semantically meaningful language field.
\item For inference, we propose an adaptive language-to-instance prompt refinement that utilizes the 3D relevance map from the language field to guide \phil{the} refinement in the instance field.

\item Our method 
\phil{sets a new}
state-of-the-art performance on two standard benchmarks and shows strong potential for image-based segmentation, hierarchical segmentation, and robotics.

\end{itemize}

\section{Related work}
\vspace{-2mm}

\paragraph{Radiance field.}
Radiance fields have emerged as a powerful representation \phil{for supporting 3D scene reconstruction} 
with diverse properties such as geometry, color, and semantics, from only 2D inputs \phil{such as multi-view} RGB images and extracted feature maps.
Neural Radiance Fields (NeRF)~\citep{mildenhall2021nerf} model the radiance field using neural networks composed of MLPs,
enabling photorealistic novel view synthesis.
Subsequent works focus on improving the efficiency of NeRF by introducing explicit 3D structures, such as voxel grids~\citep{chen2022tensorf, liu2020neural} and hash grids~\citep{muller2022instant}.
More recently, 3D Gaussian Splatting (3D-GS)~\citep{kerbl20233d, xu2024texture, liang2024analytic, zhang2024pixel, cheng2024gaussianpro, huang20242d, yu2024mip,seiskari2024gaussian,zhang2024gaussian,kulhanek2024wildgaussians} \phil{is} proposed as an alternative representation \phil{by modeling the radiance field as a set of explicit 3D Gaussian points}.
This approach supports splatting-based rendering~\citep{kopanas2021point}, which is highly efficient, significantly enhancing its potential for real-time applications.
\phil{Given these advantages, we adopt 3D-GS as the backbone representation in our 3D segmentation framework.}

\paragraph{Open-vocabulary 3D segmentation.} 
\vspace{-2mm}
Open-vocabulary 3D scene segmentation has made significant progress \phil{in recent years, empowered by}  
2D foundation vision-language models (VLMs) (\eg, CLIP~\cite{clip_method}, LSeg~\cite{li2022language}, DINO~\cite{caron2021emerging}), \phil{together} with 3D representations, ranging from point clouds to radiance fields.
Early approaches~\cite{ding2023pla,2023conceptfusion,jiang2024open,liu2023partslip,peng2023openscene,yang2024regionplc,zhang2023clip,boudjoghra2024open} project 3D point clouds to 2D views to align \phil{them} with image-based features, enabling zero-shot open-vocabulary 3D segmentation.
\rsz{However, as discussed in~\cite{engelmann2024opennerf}, point cloud representation suffers from the discrete structure and typically has a lower resolution compared to images, limiting their effectiveness and applications.}

To overcome limitations of the point cloud representation, recent methods~\cite{kerr2023lerf,qin2024langsplat,Shi_2024_CVPR,jun2025dr,ji2025fastlgs,ye2023gaussian,wu2024opengaussian,li2024instancegaussian,bhalgat2024n2f2,engelmann2024opennerf,guo2024semantic} propose distilling dense VLM features into continuous radiance field representations,
enabling high-resolution novel-view synthesis for effective feature alignment and downstream tasks.
Specifically, LeRF~\cite{kerr2023lerf} first introduced \phil{the concept of} language field distillation into NeRF via \phil{a} 2D CLIP supervision.
Besides, LangSplat~\cite{qin2024langsplat}, LEGGaussians~\cite{Shi_2024_CVPR}, Dr.Splat~\cite{jun2025dr}, and FastLGS~\cite{ji2025fastlgs} adopt 3D-GS 
as an explicit scene representation, which integrates language features, for supporting open-vocabulary 3D scene understanding.
While these models achieve promising results, the segmentation quality is still limited by the weak expressiveness of \phil{the} 
directly learned language features.
\rsz{In contrast, other methods~\cite{ye2023gaussian,wu2024opengaussian,li2024instancegaussian} tackle the task sequentially by first performing class-agnostic 3D segmentation, followed by selecting the best-matched 3D segment using language queries.}
For instance, OpenGaussian~\cite{wu2024opengaussian} and InstanceGaussian~\cite{li2024instancegaussian} \phil{propose to} align 3D segmentations with 2D segmentations produced by SAM to 
\phil{leverage} 2D CLIP features \phil{to enable subsequent text grounding}.
Besides, Gaussian Grouping~\cite{ye2023gaussian} employs a 2D vision-language grounding model (\ie, GroundingDINO~\cite{liu2024grounding}) to associate 2D grounding result with 3D segments using handcrafted 2D matching techniques.
\rsz{However, these methods suffer from error accumulation, which restricts overall performance.}
\rsz{To address 
these limitations, we introduce a novel 3D prompt-based segmentation framework that \phil{collaboratively engage both} 
segmentation \phil{cues} and language cues, throughout both the training and inference stages, to \phil{optimize for} 
high-quality open-vocabulary segmentation.}

\vspace{-3mm}
\paragraph{Promptable segmentation.}
\rsz{Promptable segmentation, which aims to generate segmentations based on input prompts \phil{by} specifying the target to be segmented within an image, was introduced by the Segment Anything Model (SAM) \cite{kirillov2023segment}.}
Its effectiveness has been widely demonstrated through a 
\phil{number} of follow-up works~\cite{li2023semantic, xiong2024efficientsam, ke2023segment, zhang2023faster, zhang2023mobilesamv2}.
To \phil{adopt it} to 3D understanding, \phil{some} recent methods~\cite{cen2023segment, choi2024click, ying2024omniseg3d, kim2024garfield} propose \phil{to learn} the discriminative instance field that encode 3D segmentation information. 
At the inference, these methods leverage user-click prompts to specify the target, \phil{facilitating the production of more} accurate segmentation results.
Though these methods achieve notable \phil{improvement}, they \phil{require additional} 
manual interaction in \phil{screen space} as prompt, 
\phil{thereby limiting} their applicability in autonomous 3D systems.

\phil{Going beyond the prior works, we}
realize a novel prompt-segmentation framework to directly support open-vocabulary 3D queries as prompts for segmentation inference through innovations that actively integrate the segmentation-aware instance field and text-aligned language field.

\section{Method}
\label{sec:method}
\begin{figure}[!tbh]
\centering
\begin{overpic}[width=1.0\textwidth]{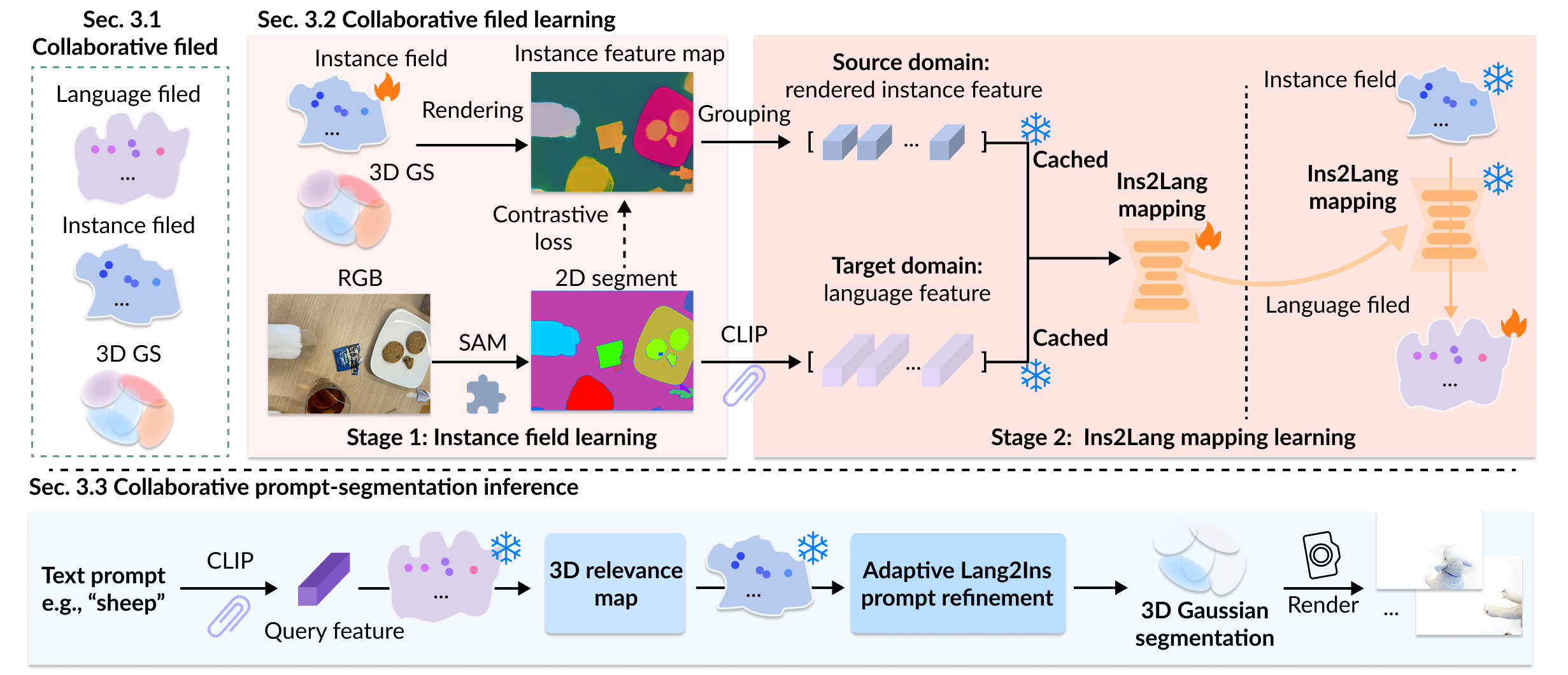}
\end{overpic}
\centering
\caption{
Overview of our proposed \ourmethod{} method. 
We first introduce collaborative field, comprising an instance field and a language field (see Sec.\ref{sec:collaborative}).
During training, we first learn the instance field to encode instance information and propose a novel instance-to-language (Ins2Lang) mapping to construct the language field (see Sec.\ref{sec:learning}).
During inference, leveraging the 3D relevance map from the language field, we design an adaptive language-to-instance (Lang2Ins) prompt refinement to further guide the instance field, enabling accurate segmentation (see Sec.\ref{sec:inference}).
}
\label{fig:overview}
\end{figure}
Given a set of multi-view posed images, we utilize the 2D foundation models SAM~\cite{kirillov2019panoptic} and CLIP~\cite{clip_method} to produce \phil{associated} 2D instance segmentation masks $\{\mathcal{K}_{I}\}$ and 2D language feature maps $\{\mathcal{F}_{L}\}$.
Based on reconstructed 3D Gaussian representations, our objective is to enrich these 3D Gaussians with high-quality, text-guided segmentations that align with the open-vocabulary query.
Fig.~\ref{fig:overview} illustrates the overview of our approach.
First, to support effective open-vocabulary segmentation, we propose the new concept of collaborative field, which comprise an instance field and a language field, integrating both segmentation and language cues (see Sec.~\ref{sec:collaborative}).
Next, to enable collaboration between the instance and language fields throughout the whole pipeline, we further design a novel instance-to-language mapping within a two-stage learning strategy in Sec.~\ref{sec:learning} and an adaptive language-to-instance prompt refinement for collaborative prompt-based inference in Sec.~\ref{sec:inference}.
\subsection{Collaborative field}
\paragraph{3D-GS backbone.}
\rsz{Specifically, our collaborative field utilize the explicit 3D Gaussian Splatting (3D-GS)~\cite{kerbl20233d}, as the underlying 3D scene representation.
}
Specifically, the 3D-GS model represents a 3D scene using a collection of explicit 3D Gaussians and leverages differentiable rasterization for efficient rendering. 
Mathematically, 3D-GS represents a 3D scene as a set of $N$ Gaussians, $G = \{g_i\}_{i=1}^N$, where each $g_i = (\mathbf{p}_i, \mathbf{s}_i, \mathbf{q}_i, o_i, \mathbf{c}_i)$ denotes the center, scale, orientation (quaternion), opacity, and color coefficients (in spherical harmonics) of the $i$-th Gaussian.
Each 3D Gaussian is projected to the image plane as a 2D Gaussian $G'_i$ via tile-based rasterization~\cite{kerbl20233d}. The color at a query pixel $u$ is computed using $\alpha$-blending:
\vspace{-0.2cm}
{\small \begin{equation}
    \mathbf{C}_u = \sum_{i \in \mathcal{N}} \mathbf{c}_i \alpha_i \prod_{t=1}^{i-1}(1 - \alpha_t), \quad \alpha_i = o_i G'_i(u),
\end{equation}
\vspace{-0.1cm}
}
where $\mathcal{N}$ is the set of sorted 2D Gaussians contributing to pixel $u$. The parameters $\{g_i\}$ are optimized via photometric reconstruction loss.
\paragraph{Collaborative field.}
To enable effective open-vocabulary 3D segmentation of the scene represented by 3D-GS, we propose the collaborative field to fully exploit the complementary strengths of segmentation and language information. 
Specifically, our collaborative field comprise two components: an instance field ${\Theta}_{\mathbf{I}}$ and a language field ${\Theta}_{\mathbf{L}}$.
Mathematically, the instance field is \ed{obtained by augmenting each Gaussian $g$ with a feature vector $\mathbf{I} \in \mathbb{R}^{d_{I}}$}, where $d_{I}$ is the feature dimension (set to 16 by default).
Besides, the language field is encoded as a high-dimensional language feature vector $\mathbf{L} \in \mathbb{R}^{d_{L}}$ for each Gaussian $g$, where $d_{L}$ is the dimension of language feature, \ie, 512 for CLIP~\cite{clip_method}.
\ed{
We refer to these representations as collaborative field because they interact during both training and inference.
During training, the instance field serves as a distinctive representation that simplifies learning the language field, allowing us to construct it more effectively and efficiently.
At inference time, the language field provides an initial 3D relevance map in response to a text query, which guides the instance field to refine and produce accurate open-vocabulary segmentation results.
}

\label{sec:collaborative}
\subsection{Collaborative field learning}
\label{sec:learning}

\rsz{In our collaborative field learning, the key idea is to introduce the instance-to-language mapping to model the intrinsic relationship of two fields.}
To this end, we propose a two-stage training strategy.
First, we learn a \rsz{segmentation-aware} instance field supervised by the 2D SAM~\cite{kirillov2023segment} segmentation.
\ed{
Next, based on the learned instance field, we construct the language field by learning a mapping function from paired instance and CLIP features extracted from multi-view images.
Once trained, this mapping function is applied to the instance field to generate a language field, completing the construction of the collaborative field.
}
\paragraph{Stage 1: instance field learning.}
For the instance field $\Theta_{I}$, we adopt contrastive learning to optimize the rendered features.
Similar to color rendering, we apply differentiable rasterization to efficiently render the instance feature $\mathbf{I}_u$ at each pixel $u$ as:
$
\mathbf{I}_{u}=\sum_{i\in\mathcal{N}}{\mathbf{I}_{i}\alpha_{i}\prod_{t=1}^{i-1}(1-\alpha_{t})}, \alpha_{i}=o_{i}G'_{i}(u).
$
Then, we apply widely-used InfoNCE loss in existing works\cite{choi2024click,ying2024omniseg3d,silva2024contrastive,zhu2025rethinking} for supervision:
\vspace{-0.1cm}
{\small
\begin{equation}
\mathcal{L}_{\text{ins}}=-\frac{1}{|\Omega|} \sum_{\Omega_{j} \in \Omega} \sum_{u \in \Omega_{j}} \log \frac{\exp \left({\operatorname{sim}}\left(\mathbf{I}_u, \overline{\mathbf{I}}_{j}\right)\right)}{\sum_{\Omega_{I} \in \Omega} \exp \left({\operatorname{sim}}\left(\mathbf{I}_{u}, \overline{\mathbf{I}}_{l}\right)\right)},
\label{eq:contra}
\end{equation}}
\vspace{-0.05cm}
where similarity kernel function $\operatorname{sim}$ uses the dot product operation here and $\Omega$ is the set of pixel samples. In specific, $\Omega_{j}, \Omega_{l}$ denotes the pixel samples with the same instance ID $j, l$ according to the 2D instance segmentation $\mathcal{K}_{I}$,
$\overline{\mathbf{I}}_j$ and $\overline{\mathbf{I}}_l$ represent the mean instance features (centroids) for $\Omega_{j}$ and $\Omega_{l}$, respectively.
\rszfinal{Notably, the instance segmentation mask $\mathcal{K}_{I}$ is automatically generated using SAM by creating a grid of point prompts across the image, following common practices~\cite{qin2024langsplat,wu2024opengaussian,jun2025dr,Shi_2024_CVPR}.}
\ed{By minimizing this loss across training views, the instance field learns to produce discriminative and view-consistent features that capture 3D instance-level information.}
\paragraph{Stage 2: instance-to-language (Ins2Lang) mapping learning.}
Based on the learned discriminative instance field, we introduce an Ins2Lang mapping to transfer the instance feature source domain to the language feature target domain.
Mathmatically, the mapping function $\Phi$ is defined as:
\begin{equation}
    \Phi:\mathbf{L} = \Phi(\mathbf{I}),\quad \mathbf{I} \in \Theta_{\mathbf{I}},
\end{equation}
where $\mathbf{I}$ denotes the instance feature and $\mathbf{L}$ denotes the corresponding language feature.
To learn the mapping function $\Phi$, we first construct training pairs between the instance features and their corresponding language features.
Specifically, for each individual view, we render the multi-view consistent instance feature map, enabling us to directly use the 2D language feature map $\mathcal{F}_{L}$ from CLIP as the corresponding pair.
Moreover, since the CLIP features are inherently patch-wise, and to reduce redundancy, we utilize the SAM mask to group rendered instance feature maps and CLIP feature maps by averaging the features with the same 2D instance ID.
This process results in segment-wise training feature pairs, denoted as ${(\mathbf{I}^m, \mathbf{L}^m)}_{m=1}^{M}$, where $M$ is the total number of pairs.
Based on the training pairs, we further provide two implementation strategies, \ie, shallow MLPs and kernel regression for the mapping function.
i) \textbf{Shallow MLPs}.
We can choose shallow MLPs, denoted as $\Phi_{\text{network}}$, to represent the instance-to-language mapping.
Then, we utilize the prepared mapping pair ${(\mathbf{I}^m, \mathbf{L}^m)}_{m=1}^{M}$ to supervise the learning of $\Phi_{\text{network}}$.
Specifically, we use the following mapping loss:
$\mathcal{L}_{\text{mapping}}= |\mathbf{L}^{m}-\Phi_{\text{network}}({\mathbf{I}}^{m})|$.
Notably, the mapping learning is highly efficient, requiring less than three minutes on a single GPU.

ii) \textbf{Kernel regression}. 
We can also utilize traditional kernel regression, denoted as $\Phi_{\text{kernel}}$, to represent the mapping function.
In particular, we adopt the widely used Nadaraya-Watson estimator~\cite{nadaraya1964estimating} because of its simplicity and learning-free nature.
Mathematically, the kernel regression function is defined as:
$\Phi_{\text{kernel}}(\mathbf{I}) = {\sum_{m=1}^M (\exp\left(-\frac{\|\mathbf{I} - {\mathbf{I}}^m\|^2}{2\sigma^2}\right) {\mathbf{L}}^{m}})/{\sum_{m=1}^M \exp\left(-\frac{\| \mathbf{I} - {\mathbf{I}}^{m}\|^2}{2\sigma^2}\right)},$
where $\sigma$ is a hyperparameter that controls the bandwidth of the kernel function, and is set to $0.1$ by default.
\rsz{Based on the learned Ins2Lang mapping $\Phi$, we can obtain our language feature field $\Theta_{L}$ by calculating the corresponding language feature $\Phi$($\mathbf{L})$ for each Gaussian.}
\paragraph{Discussion.}
\rsz{We introduce the Ins2Lang mapping within a two-stage training strategy for language field construction}, offering the advantages of effectiveness and efficiency.
1) \textbf{Effectiveness}: 
\ed{
In contrast to existing approaches that directly learn language features from scratch, our method constructs the language field via a mapping function from the learned instance field. 
This strategy leverages the discriminative, segmentation-aware features already captured by the instance field, enabling more semantically meaningful and spatially coherent language representations. 
As a result, it significantly improves the quality of open-vocabulary segmentation.
} 
2) \textbf{Efficiency}: 
\ed{
Unlike approaches that directly optimize language features for each Gaussian point, requiring extensive supervision and per-point updates, our method employs a shared mapping function that generalizes across all Gaussians. 
This significantly reduces the number of parameters and the training overhead.
Moreover, instead of using dense pixel-level supervision, we construct training pairs at the segment (patch) level using SAM masks, which further lowers redundancy and improves learning efficiency.
Together, these design choices result in a highly efficient training process.
}
In addition, the experimental comparisons with other alternatives are provided in our ablation (see Sec.~\ref{sec:ablation}).
\subsection{Collaborative prompt-segmentation inference}
\label{sec:inference}

\rsz{
During inference, we further design an adaptive language-to-instance prompt refinement, enabling a collaborative prompt-segmentation inference.}
As illustrated at the bottom of Fig.~\ref{fig:overview}, given a query text, we utilize the text-aligned language field to generate a 3D relevance map as a prompt and introduce the adaptive prompt refinement in the boundary-aware instance field for producing accurate 3D Gaussian segmentation.
\textbf{3D relevance map in language field.}
\rsz{Based on the text-aligned language field, for the text query $q_{\text{text}}$, we first generate the 3D relevance map, which indicates the correspondence between the input text and 3D regions.}
Specifically, we first utilize the CLIP~\cite{clip_method} text encoder to obtain the corresponding language feature $\mathbf{L}_{\text{text}}$.
Then, we compute the 3D point-level relevance $R$ as:
$
R = \min _i \frac{\exp \left(\mathbf{L} \cdot \mathbf{L}_{\text{text}}\right)}{\exp \left(\mathbf{L} \cdot \mathbf{L}_{\text{text}}\right)+\exp \left(\mathbf{L} \cdot \mathbf{L}_{\text{canon}}^i\right)},
$
where $\mathbf{L}_{\text{canon}}$ is the CLIP embedding of a predefined
canonical phrase~\cite{kerr2023lerf}.
Intuitively, we can obtain segmentation by identifying Gaussian points $\mathcal{S}$ with high relevance, using a predefined threshold $\tau$ (set to 0.5 by default), following common practices~\cite{kerr2023lerf,qin2024langsplat,Shi_2024_CVPR,jun2025dr, wu2024opengaussian}.

\begin{wrapfigure}[7]{r}{0.43\textwidth} 
\vspace{-1.8em}
  \begin{center}
    \includegraphics[width=0.43\textwidth]{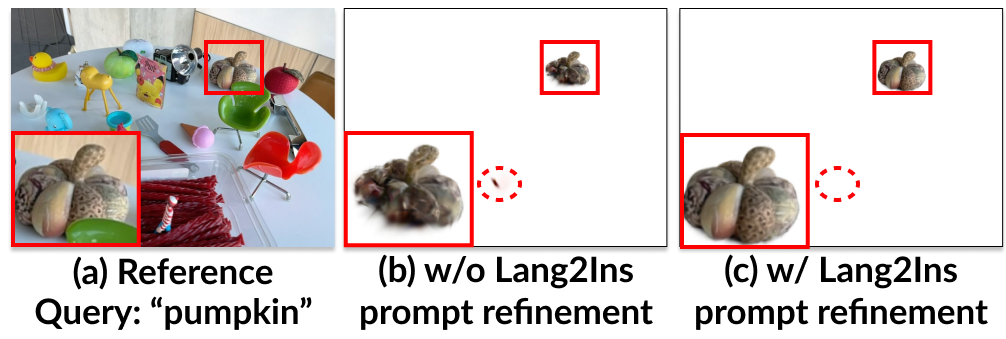}
  \end{center}
\vspace{-5mm}
  \caption{
Visual comparisons on LeRF~\cite{kerr2023lerf}.
}
\label{fig:prompt-segmentation}
\end{wrapfigure}

\textbf{Adaptive language-to-instance (Lang2Ins) prompt refinement.}
\rsz{Directly extracting the segmentation by the above process easily produces inferior segmentation (see Fig.~\ref{fig:prompt-segmentation} (b)).
To address this, we treat the relevance map as a prompt to guide the instance field in refining segmentation via an adaptive Lang2Ins prompt refinement process.}
\rsz{Particularly, starting from the initial high-relevance Gaussian point set $\mathcal{S}$, we aim to obtain a refined Gaussian point} set $\mathcal{S}_{t}$ that represents 3D Gaussian
segmentation results.
\ed{
Concretely, for each center Gaussian point $g' \in \mathcal{S}$, we leverage the learned instance field to define a local neighborhood set $\mathcal{S}_{g'}$, consisting of points whose instance features have cosine similarity above a threshold $\mathcal{T}$ with that of $g'$.
}
\rsz{Here, threshold $\mathcal{T}$ is based on the statistical value from instance field.}
Considering the presence of errors in $\mathcal{S}$, and the risk that expansion from noisy points may include undesired objects, we further perform an adaptive filtering operation for $\mathcal{S}_{g'}$ based on \ed{a} region-level relevance.
\ed{We define this region-level relevance as the opacity-weighted average of relevance scores in $\mathcal{S}_{g'}$:
}
$(\sum_{w\in \mathcal{S}_{g'}} o_{w}*R_{w})/(\sum_{w\in \mathcal{S}_{g'}}o_{w}),$ where $o_{w}$ and $R_{w}$ are the opacity and relevance of point $w \in \mathcal{S}_{g'}$.
\ed{
We process the regions in descending order based on the relevance score of their center points. 
A region is included in the final segmentation only if its region score exceeds the threshold $\tau$ (as defined earlier).
}
\rsz{By applying this process to the initial set $\mathcal{S}$,
we gradually aggregate retained expanded point sets, producing the final segmentation $\mathcal{S}_{t}$.}
More details on the algorithm and automatic \ed{threshold} generation are provided in Supp.
\textbf{Discussion.}
\rsz{Unlike the existing methods~\cite{qin2024langsplat,Shi_2024_CVPR,jun2025dr} that solely rely on the relevance map, bounded by the limited expressivity of language features, our adaptive Lang2Ins prompt refinement further leverages the discriminative instance field to aggregate neighboring points with spatial and semantic coherence, thus enabling boundary-aware segmentation.
Further, it helps adaptively filter noisy points using robust region-wise relevance.
\rsz{As shown in Fig.~\ref{fig:prompt-segmentation} (c), this algorithm significantly enhances the Gaussian segmentation quality, enabling clean object rendering.}
Moreover, the proposed adaptive Lang2Ins prompt refinement is efficient, adding small overhead to the query time (see Sec.~\ref{sec:ablation}).}
\subsection{Implementation details}
\label{sec:implementation}
We adopt the official implementation of 3D-GS~\cite{kerbl20233d} with a default of 30K training iterations as our base architecture.
For the instance field and instance-to-language mapping (\eg, MLPs version), we also set the training iterations to 30K by default, following common practice as in~\cite{qin2024langsplat}.
For kernel regression version in mapping, the function is directly formulated without requiring training. 
All experiments are conducted on a single RTX-4090 GPU.
More details are provided in Supp.

\section{Experiments}
\label{sec:exp}

\subsection{Results on LeRF dataset}
\label{sec:rendering-exp}
\paragraph{Settings.} \textbf{1) Task}: 
For open-vocabulary text queries, we first select the corresponding Gaussians and render them into multi-view 2D images.
\textbf{2) Dataset and metrics}: 
Following OpenGaussian~\cite{wu2024opengaussian} and LangSplat~\cite{qin2024langsplat}, we evaluate our method on the LeRF dataset~\cite{kerr2023lerf}. 
After rendering the selected 3D objects into 2D views, we compute mean Intersection over Union (mIoU) and mean Accuracy (mAcc). 
\textbf{3) Baselines}: 
As this task requires explicit 3D point-level segmentation, we compare our method with other explicit Gaussian-based methods, including language-based methods LangSplat~\cite{li2024langsurf}, LEGaussians~\cite{Shi_2024_CVPR}, and segmentation-based method OpenGaussian~\cite{wu2024opengaussian}.
Moreover, we provide the quantitative comparison with the most recent works, \ie, InstanceGaussian~\cite{li2024instancegaussian} and Dr. Splat~\cite{jun2025dr}. 

\begin{figure}[!ht]
\centering
\begin{overpic}[width=1.0\textwidth]{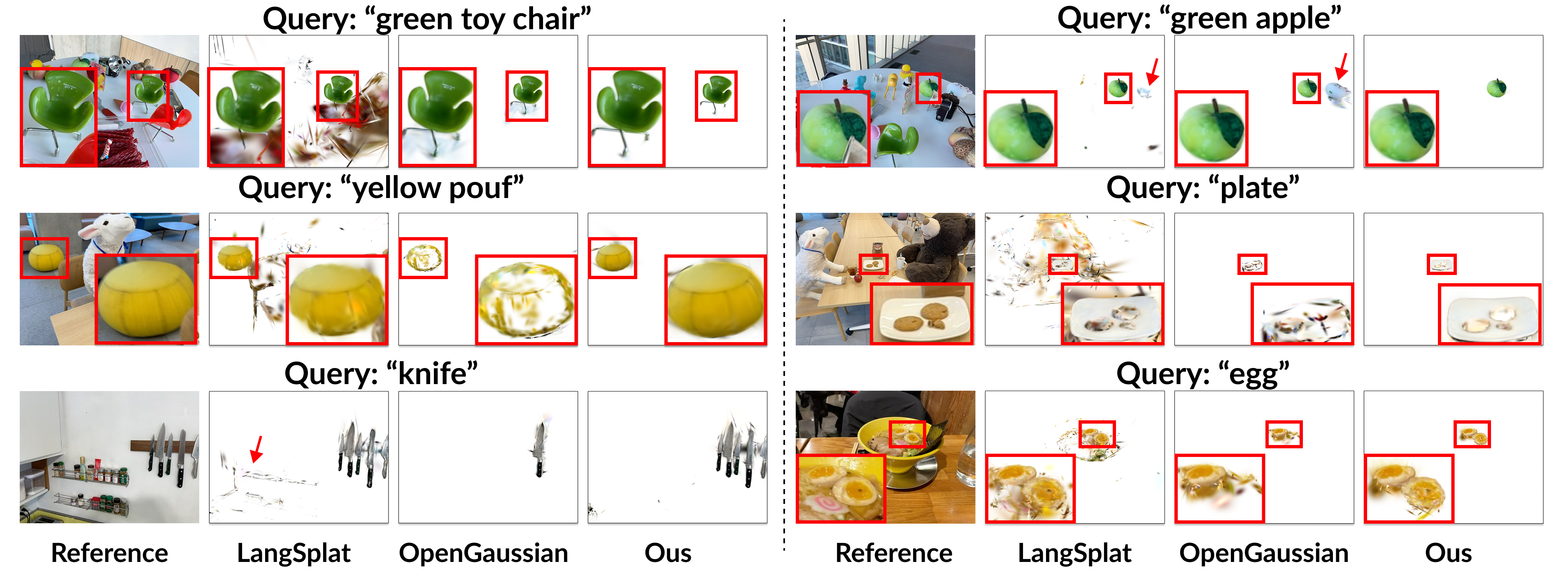}
\end{overpic}
\centering
\caption[fig:lerf]{
Open-vocabulary 3D Gaussian segmentation on the LeRF dataset. Our method outperforms
previous open-sourced SOTA methods (\ie, LangSplat, OpenGaussian) in accurately identifying the 3D objects corresponding to text queries with fewer artifacts.
Here, we present our results using kernel regression for visual comparison, and more results are provided in Supp. 
}
\label{fig:lerf}
\end{figure}

\paragraph{Results.} 
Quantitative results are presented in Tab.~\ref{tab:lerf}, demonstrating that our proposed method achieves significantly improved results compared to all existing language-based and segmentation-based baselines.
The qualitative results in Fig.~\ref{fig:lerf} further show that the rendered objects using our method contain more complete boundaries and significantly fewer artifacts.
Note that InstanceGaussian~\cite{li2024instancegaussian} and Dr. Splat~\cite{jun2025dr} are not open-sourced, which prevents further visual comparisons.
\begin{table}[!ht]
\centering
\caption{
Performance of Gaussian segmentation in 3D space from text query on LeRF~\cite{kerr2023lerf}. 
Following~\cite{wu2024opengaussian}, we report mIoU and mAcc. 
The performance of all prior works is sourced from~\cite{wu2024opengaussian,li2024instancegaussian,jun2025dr}. 
}
\resizebox{\textwidth}{!}{
\begin{tabular}{l|c|c|cc|cc|cc|cc|cc}
\toprule
& &  & \multicolumn{2}{c|}{{\textbf{mean}}} & \multicolumn{2}{c|}{\textit{figurines}} & \multicolumn{2}{c|}{\textit{teatime}} & \multicolumn{2}{c|}{\textit{ramen}} & \multicolumn{2}{c}{\textit{waldo\_kitchen}} \\
Method & Venue & Type & \textbf{mIoU} & \textbf{mAcc} & mIoU & mAcc & mIoU & mAcc & mIoU & mAcc & mIoU & mAcc \\
\midrule
LangSplat~\cite{qin2024langsplat}& CVPR'24 & Language & 9.66 & 12.41 & 10.16 & 8.93 & 11.38 & 20.34 & 7.92 & 11.27 & 9.18 & 9.09 \\
LEGaussians~\cite{Shi_2024_CVPR}& CVPR'24 & Language &  16.21 & 23.82 & 17.99 & 23.21 & 19.27 & 27.12 & 15.79 & 26.76 & 11.78 & 18.18 \\
Dr.Splat~\cite{jun2025dr} & CVPR'25 & Language  &  43.58 &  63.87 &  53.36 & \cellcolor{yellow!25}80.36 & 57.20 & 76.27 & 24.70 & 35.21 & 39.07 & 63.64 \\
\midrule
OpenGaussian~\cite{wu2024opengaussian} & NeurIPS'24 &  Segmentation  &{38.36} & 51.43 & 39.29 & 55.36 & {60.44} & 76.27 & {31.01} & {42.25} & 22.70 & 31.82 \\
IntanceGaussian~\cite{li2024instancegaussian}& CVPR'25 &  
Segmentation  & 45.30 &  58.44 & - & - & - & -& -&- & -& - \\
\midrule
Ours (shallow MLPs) & - &  Collaborative prompt &  \cellcolor{yellow!25}49.75 & \cellcolor{yellow!25}70.60 & \cellcolor{yellow!25}53.90 & 76.79 & \cellcolor{red!25}\textbf{66.91} & \cellcolor{yellow!25}88.14   & \cellcolor{red!25}\textbf{36.61} & \cellcolor{red!25}\textbf{49.30} & \cellcolor{yellow!25}41.56 & \cellcolor{red!25}68.18  
\\ 
Ours (kernel regression)& - & Collaborative prompt & \cellcolor{red!25}\textbf{50.76} & \cellcolor{red!25}\textbf{72.08} & \cellcolor{red!25}\textbf{60.03} & \cellcolor{red!25}\textbf{82.14} & \cellcolor{yellow!25}65.07 & \cellcolor{red!25}\textbf{91.53}   & \cellcolor{yellow!25}35.86 & \cellcolor{yellow!25}46.48 & \cellcolor{red!25}\textbf{42.10} & \cellcolor{red!25}\textbf{68.18}  \\ 

\bottomrule
\end{tabular}
\label{tab:lerf}
}
\vspace{-4mm}
\end{table} 

\begin{table}[!t]
\centering
\caption{Performance of Gaussian segmentation on the ScanNetv2~\cite{dai2017scannet} dataset compared to baselines~\cite{qin2024langsplat,Shi_2024_CVPR,wu2024opengaussian} based on text query.
The performance of all prior work has
been sourced from~\cite{wu2024opengaussian}. 
}
\resizebox{\textwidth}{!}{

\begin{tabular}{l|c|cccccc}
\toprule
\multirow{2}{*}{Methods} & \multirow{2}{*}{Type}& \multicolumn{2}{c}{19 classes} & \multicolumn{2}{c}{15 classes} & \multicolumn{2}{c}{10 classes} \\
             & &  mIoU $\uparrow$    & mAcc. $\uparrow$  & mIoU $\uparrow$   & mAcc. $\uparrow$  & mIoU $\uparrow$   & mAcc. $\uparrow$  \\ 
\midrule
LangSplat~\cite{qin2024langsplat}  & Language  & 3.78       & 9.11       & 5.35       & 13.20       & 8.40       & 22.06       \\
LEGaussians~\cite{Shi_2024_CVPR} & Language  & 3.84       & 10.87       & 9.01       & 22.22       & 12.82       & 28.62       \\
\midrule
OpenGaussian~\cite{wu2024opengaussian}& 
Segmentation  &24.73 & \cellcolor{yellow!25} 41.54 & 30.13 & \cellcolor{yellow!25}48.25 & \cellcolor{yellow!25}38.29 & 55.19 \\ 
\midrule
Ours (shallow MLPs) & Collaborative prompt &\cellcolor{yellow!25}26.72 & 39.89 &  \cellcolor{yellow!25}31.02 & 46.30 & 37.28 & \cellcolor{yellow!25}55.41 \\
Ours (kernel regression)  & Collaborative prompt &\cellcolor{red!25}\textbf{32.47}&\cellcolor{red!25}\textbf{49.05}&\cellcolor{red!25}\textbf{35.95}&\cellcolor{red!25}\textbf{54.35}&\cellcolor{red!25}\textbf{44.32}&\cellcolor{red!25}\textbf{63.66}\\
\bottomrule
\end{tabular}
}
\label{tab:scannet}
\end{table}
\begin{figure}[!th]
\centering
\begin{overpic}[width=1.0\textwidth]{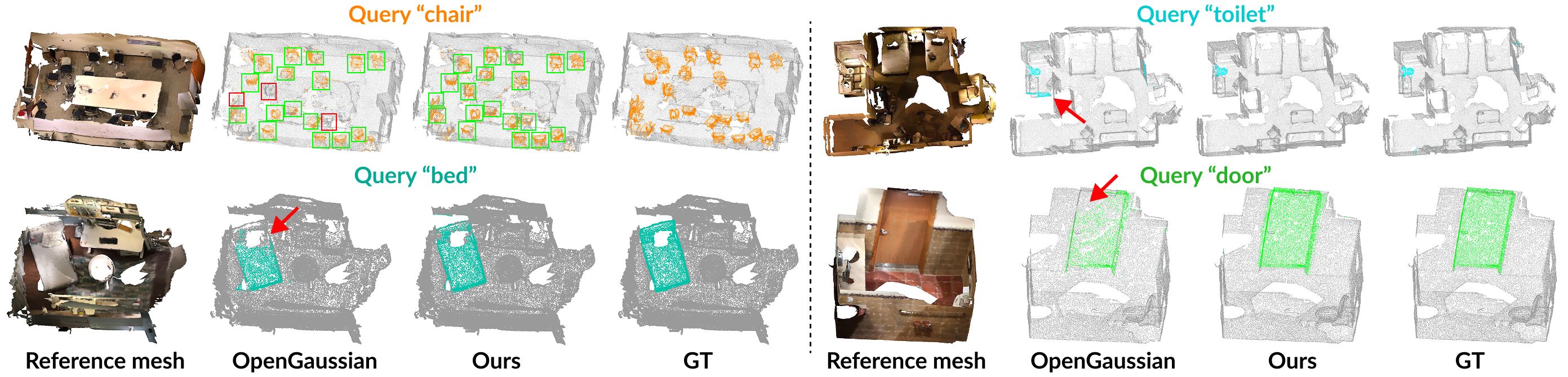}
\end{overpic}
\centering
\caption{
Open-vocabulary Gaussian segmentation on ScanNetv2~\cite{dai2017scannet} dataset. 
Our method outperforms the previous open-sourced SOTA approach (\ie, OpenGaussian) in accurately identifying 3D objects for various text queries.
In addition, we use \textcolor{green}{green} boxes to indicate regions of accurate predictions and \textcolor{red}{red} boxes to indicate regions of missing predictions.
Here, we present our results using kernel regression for visual comparison, and more results are provided in Supp. 
}
\vspace{-2mm}
\label{fig:scannet}
\end{figure}
\subsection{Results on ScanNetv2 dataset}
\label{sec:3d-exp}
\paragraph{Settings.} \textbf{1) Task}: 
In this task, we focus on direct 3D evaluation without rendering processing. 
Specifically, the model takes text queries as input and selects the corresponding Gaussian points. 
\textbf{2) Dataset and metrics}: 
Following the protocol established by OpenGaussian~\cite{wu2024opengaussian}, we adopt 19, 15, and 10 categories from ScanNetv2~\cite{dai2017scannet} as text queries. 
Moreover, we evaluate performance using mIoU and mAcc for the 10 scenes selected by OpenGaussian~\cite{wu2024opengaussian}.
\textbf{3) Baselines:} We compare our method against recent Gaussian-based approaches, including LangSplat~\cite{qin2024langsplat}, LEGaussians~\cite{Shi_2024_CVPR}, and OpenGaussian~\cite{wu2024opengaussian}, following prior work~\cite{wu2024opengaussian}.
\rszfinal{Note that the latest baselines (\ie, InstanceGaussian and Dr. Splat) use different evaluation protocols or segmentation inputs, which hinder direct evaluation using the results from their papers.
We provide additional comparisons in Supp.}
\paragraph{Results.} Quantitative results, as shown in Tab.~\ref{tab:scannet}, consistently demonstrate that our method significantly outperforms Gaussian-based methods. 
The 3D visualization results are presented in Fig.~\ref{fig:scannet}, illustrating that our method achieves accurate and complete 3D Gaussian point-level segmentation for various queries, especially in challenging scenarios (\eg, the dense ``chair'' query shown in Fig.~\ref{fig:scannet}).
\subsection{Ablation study}
\label{sec:ablation}
\paragraph{The influence of learning designs for collaborative field.}
For collaborative field learning, we 
propose an instance-to-language mapping design within a two-stage learning strategy.
\rsz{Alternatively, there are two other technically feasible training solutions for our proposed collaborative field.
One straightforward approach is the one-stage joint learning of instance and language branches, where the instance field is also influenced by the mapping loss, leading to costly optimization and an additional risk that the supervision for the mapping may make the instance feature space less discriminative.
Another alternative is parallel learning, where the two branches are trained independently, failing to fuse the instance field and the language field.}
%
%
%
%
%
%
%
As shown in Tab.~\ref{tab:ablation-mapping}, compared to alternatives, both of our two implementations for the proposed two-stage instance-to-language mapping design not only improve final performance but also significantly reduce the required training time.
Moreover, we find that our kernel regression implementation achieves the best performance.
We attribute this to the choice of discriminative instance features as the source domain, which makes the mapping process inherently an easy regression task, and the traditional kernel regression method is well-suited for such a case.
Thus, we use the kernel regression implementation for subsequent analysis.
\rszfinal{Moreover, we provide a more detailed analysis of kernel regression and the MLP counterpart in Supp.}

\paragraph{The influence of collaborative prompt-segmentation inference.}
To analyze the influence of collaborative prompt-based segmentation inference, we compare our results with two other alternatives.
Specifically, the first inference alternative conducts the class-agnostic 3D segmentation by clustering~\cite{bhalgat2023contrastive} and uses a similar strategy as OpenGaussian~\cite{wu2024opengaussian} to select the 3D segment results, and the second inference alternative solely utilizes the language branch in our collaborative field to generate the segmentation results~\cite{qin2024langsplat,Shi_2024_CVPR}.
The comparisons are presented in Tab.~\ref{tab:ablation-prompt}, indicating that collaborative prompt-based inference significantly improves segmentation by integrating language and instance knowledge, with only a slight increase in inference time per query.
\begin{table}[!htbp]
    \centering
    \begin{minipage}{0.5\textwidth}
        \centering
        \caption{Ablation of learning on LeRF~\cite{kerr2023lerf}.}
        \label{tab:ablation-mapping}
        \resizebox{\textwidth}{!}{
        \begin{tabular}{l|ccc}
\toprule
Learning solution /\ (\textbf{Mean}) & mIoU & mAcc  & Training time\\
\midrule
Joint learning &49.15 &  69.19 & 165 min \\
Parallel learning & 43.84 & 59.81 & 95 min  \\
\midrule
Our  (shallow MLPs) &   
\cellcolor{yellow!25} 49.75 & \cellcolor{yellow!25} 70.60  & 53 min 
\\ 
Our  (kernel regression)  &   \cellcolor{red!25}\textbf{50.76} & \cellcolor{red!25}\textbf{72.08} & 50 min  \\

\bottomrule
\end{tabular}
}
    \end{minipage}
    \begin{minipage}{0.48\textwidth}
        \centering
        \caption{Ablation of inferences on LeRF~\cite{kerr2023lerf}.}
        \label{tab:ablation-prompt}
        \resizebox{\textwidth}{!}{
              \begin{tabular}{l|ccc}
\toprule
Inference solution /\ (\textbf{Mean}) & mIoU & mAcc & Query time  \\
\midrule
Instance branch  &   44.07 & 59.83 & 0.12 s \\ 
Language branch  &   \cellcolor{yellow!25}48.99 & \cellcolor{yellow!25}71.31 & 0.13 s\\
\midrule
Collaborative prompt (ours) &   \cellcolor{red!25}\textbf{50.76} & \cellcolor{red!25}\textbf{72.08} & 0.22 s \\
\bottomrule
\end{tabular}
        
        }
    \end{minipage}
\end{table}
\paragraph{Training efficiency.}
To analyze the training efficiency of our method, we compare performance under different training times.
As training the instance field with the default 30K optimization steps requires the majority of the total training time (45 out of 50 minutes), we conducted experiments with shorter training times by reducing the number of optimization steps (\ie, 3K and 6K).
The results, presented in Tab.~\ref{tab:time}, demonstrate that our method converges quickly and achieves significantly better performance than baselines even with less training time, highlighting our superior training efficiency.

\paragraph{\rszfinal{The influence of different 2D foundation vision-language models (VLMs).}}
We utilize CLIP~\cite{clip_method} and SAM~\cite{kirillov2023segment} as the default 2D language and segmentation models in our main experiments, following common practice in recent baselines~\cite{qin2024langsplat,Shi_2024_CVPR,wu2024opengaussian,jun2025dr,li2024instancegaussian}, to ensure that the improvements are attributed to the proposed collaborative fields design.
Furthermore, we conducted an ablation study comparing different 2D foundation VLMs (e.g., CLIP~\cite{clip_method} vs. SigLIP~\cite{zhai2023sigmoid}, SAM~\cite{kirillov2023segment} vs. SAM2~\cite{ravi2024sam}, Semantic SAM~\cite{li2023semantic}). 
The results, shown in the Tab.~\ref{tab:vlm}, demonstrate that our framework is compatible with different 2D foundation models.
Additionally, we empirically observed that using more advanced models (\eg, SAM2~\cite{ravi2024sam} for the segmentation model and SigLIP~\cite{zhai2023sigmoid} for the language model) can lead to performance improvements. 
\begin{table}[!htbp]
    \centering
    \begin{minipage}{0.43\textwidth}
        \centering
        \caption{Training efficiency analysis on LeRF~\cite{kerr2023lerf}.
  We report mean mIoU.
  } 
        \label{tab:time}
        \resizebox{\textwidth}{!}{
 \begin{tabular}{lcc}
  \toprule
    Method & Training time & mIoU  \\
    \midrule
    Langsplat~\cite{qin2024langsplat} & 240 min & 9.66 \\
    LEGaussian~\cite{Shi_2024_CVPR} & 240 min & 16.21 \\
    Dr.Splat~\cite{jun2025dr} & 10 min& 43.58 \\
    OpenGaussian~\cite{wu2024opengaussian} & 60 min & 38.36 \\
    Instance Gaussian~\cite{li2024instancegaussian} & - & 45.30 \\
    \hline
    Ours (3k for instance) & 8 min & 50.16\\
    Ours (6k for instance) & 15 min & 50.24\\
    Ours (default) & 50 min & 50.76 \\
    \bottomrule
  \end{tabular}
  }    
    \end{minipage}
    \begin{minipage}{0.55\textwidth}
        \centering
        \caption{Comparisons of various 2D foundation VLMs on LeRF~\cite{kerr2023lerf}. Model A is the default implementation for our proposed COS3D.}
        \label{tab:vlm}
        \resizebox{\textwidth}{!}{
\begin{tabular}{lcccc}
\toprule
Model & Segmentation & Language & mIoU  & mAcc   \\
\midrule
A & SAM~\cite{kirillov2023segment} & CLIP~\cite{clip_method} & 50.76 & 72.08 \\
B & SAM~\cite{kirillov2023segment} & SigLIP~\cite{zhai2023sigmoid} & 51.08 & 73.79 \\
C & SAM2~\cite{ravi2024sam} & CLIP~\cite{clip_method} & 51.94 & 75.05 \\
D & Semantic SAM~\cite{li2023semantic} & CLIP~\cite{clip_method} & 49.93 & 70.94 \\
\bottomrule
\end{tabular}
}
    \end{minipage}
\end{table}

\begin{figure}[!tbh]
\centering
\begin{overpic}[width=1.0\textwidth]{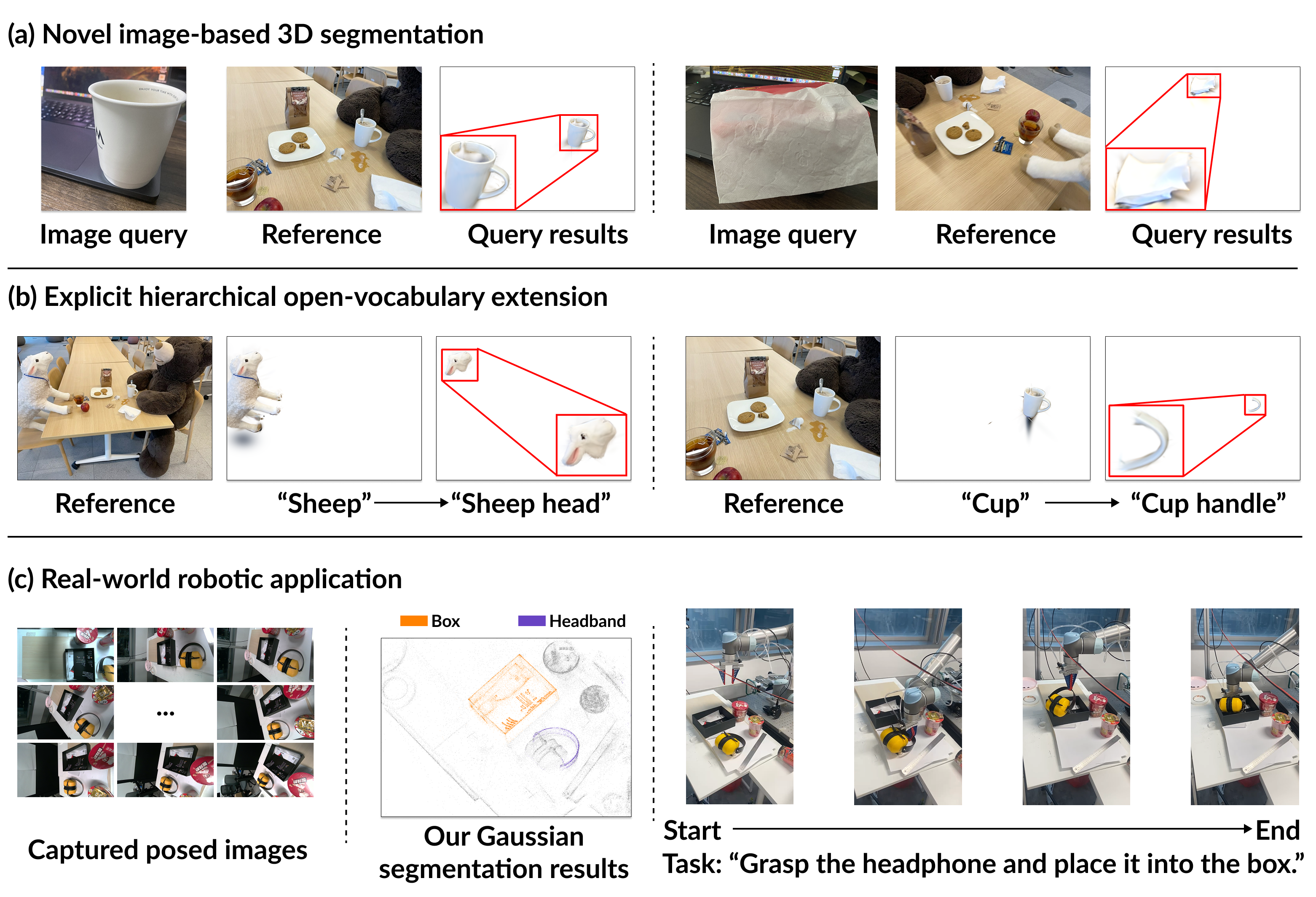}
\end{overpic}
\centering
\caption[fig:hierarchical]{
(1) We illustrate the novel image-based 3D segmentation results.
(b) We perform explicit hierarchical open-vocabulary 3D segmentation on LERF~\cite{kerr2023lerf}.
(c) 
Following prior works~\cite{zheng2024gaussiangrasper,tang2025geomanip,rashid2023language}, we leverage our method to provide accurate 3D segmentation for real-world robotic grasping, enabling the successful execution of grasp manipulation.
We provide the video demo in Supp.
}
\label{fig:hierarchica}
\end{figure}
\subsection{Applications}
\label{sec:application}
\paragraph{Novel image-based 3D segmentation.}
Beyond text queries, our method inherently supports 3D segmentation using a novel image as queries.
Specifically, given a novel query image, we utilize the CLIP vision backbone to extract visual features and apply our inference algorithm to obtain the 3D segmentation.
As shown in Fig.~\ref{fig:hierarchica} (a), our method enables accurate image-based segmentation when the query image contains a similar, but not identical, object to those in the original 3D scene.
\paragraph{Explicit hierarchical open-vocabulary extension.}
%
%
%
%
%
Furthermore, our method can be naturally extended to support explicit hierarchical queries.
Inspired by Click-Gaussian~\cite{choi2024click}, we apply our mapping function to construct a hierarchical language field based on a two-level feature representation to capture hierarchical information.
As shown in Fig.~\ref{fig:hierarchica} (b), our approach produces accurate 3D segmentation results across coarse and fine-grained queries, enabling explicit hierarchical understanding.

\paragraph{Real-world robotic application.}
We further demonstrate the applicability of \ourmethod{} in real-world robotic scenarios.
Specifically, we leverage our method to provide effective 3D perception for robotic grasping tasks, building upon prior work~\cite{zheng2024gaussiangrasper,tang2025geomanip,rashid2023language}. As illustrated in Fig.~\ref{fig:hierarchica} (c), the accurate open-vocabulary segmentation produced by our method assists the robotic arm in completing grasping operations. More details and the video demo are provided in Supp.
\section{Conclusion}
\label{sec:conclusion}
We presented \ourmethod{}, a new collaborative 3D prompt-segmentation approach for open-vocabulary queries.
We introduce the new concept of collaborative field comprising the instance and language fields.
To achieve collaboration, we model the instance-to-language mapping during training and design an adaptive language-to-instance prompt refinement during inference.
Extensive experimental results manifest the effectiveness of our method over the state of the art.

\paragraph{\rszfinal{Limitation.}}
Although \ourmethod{} provides effective open-vocabulary 3D segmentation, it has the following limitations.
First, \ourmethod{} lacks reasoning capabilities for 3D segmentation, as the text-aligned language field struggles with relational or multi-object queries.
Moreover, following recent approaches~\cite{qin2024langsplat,wu2024opengaussian,jun2025dr,Shi_2024_CVPR}, \ourmethod{} adopts the offline setting, and extending \ourmethod{} to the online setting would be beneficial.
We provide more discussion on these potential extensions in Supp.
%
%



\section*{Acknowledgements}
This study was funded by the InnoHK initiative of the Innovation and Technology Commission of the Hong Kong Special Administrative Region Government via the Hong Kong Centre for Logistics Robotics; and Hong Kong Innovation and Technology Fund under Project MHP/092/22.

{
    \small
    \bibliographystyle{ieeetr}
    \bibliography{main}
}





\appendix



\newpage
\section*{NeurIPS Paper Checklist}

 

\begin{enumerate}

\item {\bf Claims}
    \item[] Question: Do the main claims made in the abstract and introduction accurately reflect the paper's contributions and scope?
    \item[] Answer: \answerYes{} 
    \item[] Justification: We clarify our contributions in Sec.~\ref{sec:intro}, and experimentally verify them in Sec.~\ref{sec:exp}.
    \item[] Guidelines:
    \begin{itemize}
        \item The answer NA means that the abstract and introduction do not include the claims made in the paper.
        \item The abstract and/or introduction should clearly state the claims made, including the contributions made in the paper and important assumptions and limitations. A No or NA answer to this question will not be perceived well by the reviewers. 
        \item The claims made should match theoretical and experimental results, and reflect how much the results can be expected to generalize to other settings. 
        \item It is fine to include aspirational goals as motivation as long as it is clear that these goals are not attained by the paper. 
    \end{itemize}

\item {\bf Limitations}
    \item[] Question: Does the paper discuss the limitations of the work performed by the authors?
    \item[] Answer: \answerYes{} 
    \item[] Justification: We discuss the current limitations regarding 3D reasoning segmentation for handling complicated queries in Sec.~\ref{sec:conclusion}.
    \item[] Guidelines:
    \begin{itemize}
        \item The answer NA means that the paper has no limitation while the answer No means that the paper has limitations, but those are not discussed in the paper. 
        \item The authors are encouraged to create a separate "Limitations" section in their paper.
        \item The paper should point out any strong assumptions and how robust the results are to violations of these assumptions (e.g., independence assumptions, noiseless settings, model well-specification, asymptotic approximations only holding locally). The authors should reflect on how these assumptions might be violated in practice and what the implications would be.
        \item The authors should reflect on the scope of the claims made, e.g., if the approach was only tested on a few datasets or with a few runs. In general, empirical results often depend on implicit assumptions, which should be articulated.
        \item The authors should reflect on the factors that influence the performance of the approach. For example, a facial recognition algorithm may perform poorly when image resolution is low or images are taken in low lighting. Or a speech-to-text system might not be used reliably to provide closed captions for online lectures because it fails to handle technical jargon.
        \item The authors should discuss the computational efficiency of the proposed algorithms and how they scale with dataset size.
        \item If applicable, the authors should discuss possible limitations of their approach to address problems of privacy and fairness.
        \item While the authors might fear that complete honesty about limitations might be used by reviewers as grounds for rejection, a worse outcome might be that reviewers discover limitations that aren't acknowledged in the paper. The authors should use their best judgment and recognize that individual actions in favor of transparency play an important role in developing norms that preserve the integrity of the community. Reviewers will be specifically instructed to not penalize honesty concerning limitations.
    \end{itemize}

\item {\bf Theory assumptions and proofs}
    \item[] Question: For each theoretical result, does the paper provide the full set of assumptions and a complete (and correct) proof?
    \item[] Answer: \answerNA{}
    \item[] Justification: This paper does not involve theoretical results.
    \item[] Guidelines:
    \begin{itemize}
        \item The answer NA means that the paper does not include theoretical results. 
        \item All the theorems, formulas, and proofs in the paper should be numbered and cross-referenced.
        \item All assumptions should be clearly stated or referenced in the statement of any theorems.
        \item The proofs can either appear in the main paper or the supplemental material, but if they appear in the supplemental material, the authors are encouraged to provide a short proof sketch to provide intuition. 
        \item Inversely, any informal proof provided in the core of the paper should be complemented by formal proofs provided in appendix or supplemental material.
        \item Theorems and Lemmas that the proof relies upon should be properly referenced. 
    \end{itemize}
    

    \item {\bf Experimental result reproducibility}
    \item[] Question: Does the paper fully disclose all the information needed to reproduce the main experimental results of the paper to the extent that it affects the main claims and/or conclusions of the paper (regardless of whether the code and data are provided or not)?
    \item[] Answer: \answerYes{}{} 
    \item[] Justification: We disclose the experimental information in Sec.~\ref{sec:implementation}, Sec.~\ref{sec:rendering-exp}, Sec.~\ref{sec:3d-exp}, and the supp. 
    \item[] Guidelines:
    \begin{itemize}
        \item The answer NA means that the paper does not include experiments.
        \item If the paper includes experiments, a No answer to this question will not be perceived well by the reviewers: Making the paper reproducible is important, regardless of whether the code and data are provided or not.
        \item If the contribution is a dataset and/or model, the authors should describe the steps taken to make their results reproducible or verifiable. 
        \item Depending on the contribution, reproducibility can be accomplished in various ways. For example, if the contribution is a novel architecture, describing the architecture fully might suffice, or if the contribution is a specific model and empirical evaluation, it may be necessary to either make it possible for others to replicate the model with the same dataset, or provide access to the model. In general. releasing code and data is often one good way to accomplish this, but reproducibility can also be provided via detailed instructions for how to replicate the results, access to a hosted model (e.g., in the case of a large language model), releasing of a model checkpoint, or other means that are appropriate to the research performed.
        \item While NeurIPS does not require releasing code, the conference does require all submissions to provide some reasonable avenue for reproducibility, which may depend on the nature of the contribution. For example
        \begin{enumerate}
            \item If the contribution is primarily a new algorithm, the paper should make it clear how to reproduce that algorithm.
            \item If the contribution is primarily a new model architecture, the paper should describe the architecture clearly and fully.
            \item If the contribution is a new model (e.g., a large language model), then there should either be a way to access this model for reproducing the results or a way to reproduce the model (e.g., with an open-source dataset or instructions for how to construct the dataset).
            \item We recognize that reproducibility may be tricky in some cases, in which case authors are welcome to describe the particular way they provide for reproducibility. In the case of closed-source models, it may be that access to the model is limited in some way (e.g., to registered users), but it should be possible for other researchers to have some path to reproducing or verifying the results.
        \end{enumerate}
    \end{itemize}

\item {\bf Open access to data and code}
    \item[] Question: Does the paper provide open access to the data and code, with sufficient instructions to faithfully reproduce the main experimental results, as described in supplemental material?
    \item[] Answer: \answerYes{} 
    \item[] Justification: Both data and code of our work will be publicly available on github.
    \item[] Guidelines:
    \begin{itemize}
        \item The answer NA means that paper does not include experiments requiring code.
        \item Please see the NeurIPS code and data submission guidelines (\url{https://nips.cc/public/guides/CodeSubmissionPolicy}) for more details.
        \item While we encourage the release of code and data, we understand that this might not be possible, so “No” is an acceptable answer. Papers cannot be rejected simply for not including code, unless this is central to the contribution (e.g., for a new open-source benchmark).
        \item The instructions should contain the exact command and environment needed to run to reproduce the results. See the NeurIPS code and data submission guidelines (\url{https://nips.cc/public/guides/CodeSubmissionPolicy}) for more details.
        \item The authors should provide instructions on data access and preparation, including how to access the raw data, preprocessed data, intermediate data, and generated data, etc.
        \item The authors should provide scripts to reproduce all experimental results for the new proposed method and baselines. If only a subset of experiments are reproducible, they should state which ones are omitted from the script and why.
        \item At submission time, to preserve anonymity, the authors should release anonymized versions (if applicable).
        \item Providing as much information as possible in supplemental material (appended to the paper) is recommended, but including URLs to data and code is permitted.
    \end{itemize}

\item {\bf Experimental setting/details}
    \item[] Question: Does the paper specify all the training and test details (e.g., data splits, hyperparameters, how they were chosen, type of optimizer, etc.) necessary to understand the results?
    \item[] Answer: \answerYes{} 
    \item[] Justification: The experimental settings and details are presented in Sec.~\ref{sec:rendering-exp}, Sec.~\ref{sec:3d-exp}, and the supp.
    \item[] Guidelines:
    \begin{itemize}
        \item The answer NA means that the paper does not include experiments.
        \item The experimental setting should be presented in the core of the paper to a level of detail that is necessary to appreciate the results and make sense of them.
        \item The full details can be provided either with the code, in appendix, or as supplemental material.
    \end{itemize}

\item {\bf Experiment statistical significance}
    \item[] Question: Does the paper report error bars suitably and correctly defined or other appropriate information about the statistical significance of the experiments?
    \item[] Answer: \answerNo{} 
    \item[] Justification: We follow the common practice of existing works to report the  experimental results, particularly OpenGaussian (published in NeurIPS 2024)
    \item[] Guidelines:
    \begin{itemize}
        \item The answer NA means that the paper does not include experiments.
        \item The authors should answer "Yes" if the results are accompanied by error bars, confidence intervals, or statistical significance tests, at least for the experiments that support the main claims of the paper.
        \item The factors of variability that the error bars are capturing should be clearly stated (for example, train/test split, initialization, random drawing of some parameter, or overall run with given experimental conditions).
        \item The method for calculating the error bars should be explained (closed form formula, call to a library function, bootstrap, etc.)
        \item The assumptions made should be given (e.g., Normally distributed errors).
        \item It should be clear whether the error bar is the standard deviation or the standard error of the mean.
        \item It is OK to report 1-sigma error bars, but one should state it. The authors should preferably report a 2-sigma error bar than state that they have a 96\% CI, if the hypothesis of Normality of errors is not verified.
        \item For asymmetric distributions, the authors should be careful not to show in tables or figures symmetric error bars that would yield results that are out of range (e.g. negative error rates).
        \item If error bars are reported in tables or plots, The authors should explain in the text how they were calculated and reference the corresponding figures or tables in the text.
    \end{itemize}

\item {\bf Experiments compute resources}
    \item[] Question: For each experiment, does the paper provide sufficient information on the computer resources (type of compute workers, memory, time of execution) needed to reproduce the experiments?
    \item[] Answer: \answerYes{} 
    \item[] Justification: We provide machine information in Sec.~\ref{sec:implementation}. We provide the training time and inference time in the ablation study Sec.~\ref{sec:ablation}.

    \item[] Guidelines:
    \begin{itemize}
        \item The answer NA means that the paper does not include experiments.
        \item The paper should indicate the type of compute workers CPU or GPU, internal cluster, or cloud provider, including relevant memory and storage.
        \item The paper should provide the amount of compute required for each of the individual experimental runs as well as estimate the total compute. 
        \item The paper should disclose whether the full research project required more compute than the experiments reported in the paper (e.g., preliminary or failed experiments that didn't make it into the paper). 
    \end{itemize}

\item {\bf Code of ethics}
    \item[] Question: Does the research conducted in the paper conform, in every respect, with the NeurIPS Code of Ethics \url{https://neurips.cc/public/EthicsGuidelines}?
    \item[] Answer: \answerYes{} 
    \item[] Justification: We have already confirmed with the NeurIPS Code of Ethics.

    \item[] Guidelines:
    \begin{itemize}
        \item The answer NA means that the authors have not reviewed the NeurIPS Code of Ethics.
        \item If the authors answer No, they should explain the special circumstances that require a deviation from the Code of Ethics.
        \item The authors should make sure to preserve anonymity (e.g., if there is a special consideration due to laws or regulations in their jurisdiction).
    \end{itemize}

\item {\bf Broader impacts}
    \item[] Question: Does the paper discuss both potential positive societal impacts and negative societal impacts of the work performed?
    \item[] Answer: \answerYes{} 
    \item[] Justification: We discuss possible societal impacts of it in Sec.~\ref{sec:application} about its robitics applications.
    \item[] Guidelines:
    \begin{itemize}
        \item The answer NA means that there is no societal impact of the work performed.
        \item If the authors answer NA or No, they should explain why their work has no societal impact or why the paper does not address societal impact.
        \item Examples of negative societal impacts include potential malicious or unintended uses (e.g., disinformation, generating fake profiles, surveillance), fairness considerations (e.g., deployment of technologies that could make decisions that unfairly impact specific groups), privacy considerations, and security considerations.
        \item The conference expects that many papers will be foundational research and not tied to particular applications, let alone deployments. However, if there is a direct path to any negative applications, the authors should point it out. For example, it is legitimate to point out that an improvement in the quality of generative models could be used to generate deepfakes for disinformation. On the other hand, it is not needed to point out that a generic algorithm for optimizing neural networks could enable people to train models that generate Deepfakes faster.
        \item The authors should consider possible harms that could arise when the technology is being used as intended and functioning correctly, harms that could arise when the technology is being used as intended but gives incorrect results, and harms following from (intentional or unintentional) misuse of the technology.
        \item If there are negative societal impacts, the authors could also discuss possible mitigation strategies (e.g., gated release of models, providing defenses in addition to attacks, mechanisms for monitoring misuse, mechanisms to monitor how a system learns from feedback over time, improving the efficiency and accessibility of ML).
    \end{itemize}

\item {\bf Safeguards}
    \item[] Question: Does the paper describe safeguards that have been put in place for responsible release of data or models that have a high risk for misuse (e.g., pretrained language models, image generators, or scraped datasets)?
    \item[] Answer: \answerNA{} 
    \item[] Justification: The paper does not involve the release of data or models that pose a high risk for misuse.
    \item[] Guidelines:
    \begin{itemize}
        \item The answer NA means that the paper poses no such risks.
        \item Released models that have a high risk for misuse or dual-use should be released with necessary safeguards to allow for controlled use of the model, for example by requiring that users adhere to usage guidelines or restrictions to access the model or implementing safety filters. 
        \item Datasets that have been scraped from the Internet could pose safety risks. The authors should describe how they avoided releasing unsafe images.
        \item We recognize that providing effective safeguards is challenging, and many papers do not require this, but we encourage authors to take this into account and make a best faith effort.
    \end{itemize}

\item {\bf Licenses for existing assets}
    \item[] Question: Are the creators or original owners of assets (e.g., code, data, models), used in the paper, properly credited and are the license and terms of use explicitly mentioned and properly respected?
    \item[] Answer: \answerYes{} 
    \item[] Justification: We apply LeRF and ScanNetv2 dataset and 3D-GS, OpenGaussian codebase for our method implementation and testing, all of them are publicly available and have already used in published papers.
    \item[] Guidelines:
    \begin{itemize}
        \item The answer NA means that the paper does not use existing assets.
        \item The authors should cite the original paper that produced the code package or dataset.
        \item The authors should state which version of the asset is used and, if possible, include a URL.
        \item The name of the license (e.g., CC-BY 4.0) should be included for each asset.
        \item For scraped data from a particular source (e.g., website), the copyright and terms of service of that source should be provided.
        \item If assets are released, the license, copyright information, and terms of use in the package should be provided. For popular datasets, \url{paperswithcode.com/datasets} has curated licenses for some datasets. Their licensing guide can help determine the license of a dataset.
        \item For existing datasets that are re-packaged, both the original license and the license of the derived asset (if it has changed) should be provided.
        \item If this information is not available online, the authors are encouraged to reach out to the asset's creators.
    \end{itemize}


\item {\bf New assets}
    \item[] Question: Are new assets introduced in the paper well documented and is the documentation provided alongside the assets?
    \item[] Answer: \answerNA{} 
    \item[] Justification: The paper does not release new assets.
    \item[] Guidelines:
    \begin{itemize}
        \item The answer NA means that the paper does not release new assets.
        \item Researchers should communicate the details of the dataset/code/model as part of their submissions via structured templates. This includes details about training, license, limitations, etc. 
        \item The paper should discuss whether and how consent was obtained from people whose asset is used.
        \item At submission time, remember to anonymize your assets (if applicable). You can either create an anonymized URL or include an anonymized zip file.
    \end{itemize}

\item {\bf Crowdsourcing and research with human subjects}
    \item[] Question: For crowdsourcing experiments and research with human subjects, does the paper include the full text of instructions given to participants and screenshots, if applicable, as well as details about compensation (if any)? 
    \item[] Answer: \answerNA{} 
    \item[] Justification: The paper does not involve crowdsourcing nor research with human subjects.
    \item[] Guidelines:
    \begin{itemize}
        \item The answer NA means that the paper does not involve crowdsourcing nor research with human subjects.
        \item Including this information in the supplemental material is fine, but if the main contribution of the paper involves human subjects, then as much detail as possible should be included in the main paper. 
        \item According to the NeurIPS Code of Ethics, workers involved in data collection, curation, or other labor should be paid at least the minimum wage in the country of the data collector. 
    \end{itemize}


\item {\bf Institutional review board (IRB) approvals or equivalent for research with human subjects}
    \item[] Question: Does the paper describe potential risks incurred by study participants, whether such risks were disclosed to the subjects, and whether Institutional Review Board (IRB) approvals (or an equivalent approval/review based on the requirements of your country or institution) were obtained?
    \item[] Answer: \answerNA{} 
    \item[] Justification: The paper does not involve crowdsourcing nor research with human subjects.
    \item[] Guidelines:
    \begin{itemize}
        \item The answer NA means that the paper does not involve crowdsourcing nor research with human subjects.
        \item Depending on the country in which research is conducted, IRB approval (or equivalent) may be required for any human subjects research. If you obtained IRB approval, you should clearly state this in the paper. 
        \item We recognize that the procedures for this may vary significantly between institutions and locations, and we expect authors to adhere to the NeurIPS Code of Ethics and the guidelines for their institution. 
        \item For initial submissions, do not include any information that would break anonymity (if applicable), such as the institution conducting the review.
    \end{itemize}

\item {\bf Declaration of LLM usage}
    \item[] Question: Does the paper describe the usage of LLMs if it is an important, original, or non-standard component of the core methods in this research? Note that if the LLM is used only for writing, editing, or formatting purposes and does not impact the core methodology, scientific rigorousness, or originality of the research, declaration is not required.
    \item[] Answer: \answerNA{} 
    \item[] Justification: The core method development in this research does not involve LLMs as any important, original, or non-standard components.
    \item[] Guidelines:
    \begin{itemize}
        \item The answer NA means that the core method development in this research does not involve LLMs as any important, original, or non-standard components.
        \item Please refer to our LLM policy (\url{https://neurips.cc/Conferences/2025/LLM}) for what should or should not be described.
    \end{itemize}

\end{enumerate}

\end{document}